\documentclass[sn-standardnature,iicol]{sn-jnl}

\usepackage[switch]{lineno}

\jyear{2021}%

\theoremstyle{thmstyleone}%
\theoremstyle{thmstyletwo}%

\theoremstyle{thmstylethree}%

\begin{document}

\title[Article Title]{Decision-making and control with diffractive optical networks}


\author[1]{\fnm{Jumin} \sur{Qiu}}\email{qiujumin@email.ncu.edu.cn}

\author[4,5]{\fnm{Shuyuan} \sur{Xiao}}\email{syxiao@ncu.edu.cn}

\author*[2]{\fnm{Lujun} \sur{Huang}}\email{ljhuang@phy.ecnu.edu.cn}

\author[3]{\fnm{Andrey} \sur{Miroshnichenko}}\email{andrey.miroshnichenko@unsw.edu.au}

\author[1]{\fnm{Dejian} \sur{Zhang}}\email{dejianzhang@ncu.edu.cn}

\author*[4,5]{\fnm{Tingting} \sur{Liu}}\email{ttliu@ncu.edu.cn}

\author*[1]{\fnm{Tianbao} \sur{Yu}}\email{yutianbao@ncu.edu.cn}

\affil[1]{\orgdiv{School of Physics and Materials Science}, \orgname{Nanchang University}, \orgaddress{\city{Nanchang}, \postcode{330031}, \state{Jiangxi}, \country{China}}}

\affil[2]{\orgdiv{School of Physics and Electronic Science}, \orgname{East China Normal University}, \orgaddress{\city{Shanghai}, \postcode{200241}, \country{China}}}

\affil[3]{\orgdiv{School of Engineering and Information Technology}, \orgname{University of New South Wales Canberra}, \orgaddress{\city{Canberra}, \postcode{ACT 2600}, \country{Australia}}}

\affil[4]{\orgdiv{Institute for Advanced Study}, \orgname{Nanchang University}, \orgaddress{\city{Nanchang}, \postcode{330031}, \state{Jiangxi}, \country{China}}}

\affil[5]{\orgdiv{Jiangxi Key Laboratory for Microscale Interdisciplinary Study}, \orgname{Nanchang University}, \orgaddress{\city{Nanchang}, \postcode{330031}, \state{Jiangxi}, \country{China}}}


\abstract{The ultimate goal of artificial intelligence is to mimic the human brain to perform decision-making and control directly from high-dimensional sensory input. Diffractive optical networks provide a promising solution for implementing artificial intelligence with high-speed and low-power consumption. Most of the reported diffractive optical networks focus on single or multiple tasks that do not involve environmental interaction, such as object recognition and image classification. In contrast, the networks capable of performing decision-making and control have not yet been developed to our knowledge. Here, we propose using deep reinforcement learning to implement diffractive optical networks that imitate human-level decision-making and control capability. Such networks taking advantage of a residual architecture, allow for finding optimal control policies through interaction with the environment and can be readily implemented with existing optical devices. The superior performance of these networks is verified by engaging three types of classic games, Tic-Tac-Toe, Super Mario Bros., and Car Racing.
Finally, we present an experimental demonstration of playing Tic-Tac-Toe by leveraging diffractive optical networks based on a spatial light modulator. Our work represents a solid step forward in advancing diffractive optical networks, which promises a fundamental shift from the target-driven control of a pre-designed state for simple recognition or classification tasks to the high-level sensory capability of artificial intelligence. It may find exciting applications in autonomous driving, intelligent robots, and intelligent manufacturing.}

\maketitle
\clearpage

\section{Introduction}\label{sec1}

Artificial intelligence (AI) is to imitate the functions of neurons in performing decision-making by creating hierarchical artificial neural networks. It has found many exciting applications in computer vision\cite{10.1145/3065386,Russakovsky2015}, natural language processing\cite{chen-etal-2017-enhanced,devlin-etal-2019-bert}, and data mining\cite{10.1145/2939672.2939754}. Except for electronics and computer science applications, artificial neural networks have been applied to optimize the design of photonic devices, including metamaterials and metasurface, significantly facilitating the performance of photonic devices beyond the conventional inverse design strategy\cite{Ma2021,https://doi.org/10.1002/adom.202200097,https://doi.org/10.1002/adfm.202101748,Huang20,Nadell:19,10.1117/1.AP.2.2.026003,doi:10.1021/acs.nanolett.9b03971,Dai:21}. 

Recently, optical neural networks have drawn tremendous attention because they provide a compelling route of processing information at the speed of light\cite{Shen2017,Feldmann2019,PhysRevX.9.021032,Zhang2021,Liu2021,Wu:20}, with low energy consumption and massive parallelism compared to the electronic-circuit-based neural networks. In the pioneering work of Lin et al.\cite{doi:10.1126/science.aat8084}, diffractive optical networks (DON, also known as diffractive deep neural network, D$^2$NN) consisting of multilayer of three-dimensional printed diffractive optical elements operating at terahertz were first proposed for inference and prediction through parallel computation and dense interconnection at the speed of light. Later, DONs were extended to various nanostructures for implementation. Such architecture has been effectively validated in performing specific inference functions, such as image classification\cite{CHEN20211483,Luo2022,Liu2022,doi:10.1126/sciadv.abo6410}, saliency detection\cite{PhysRevLett.123.023901}, and logic operation\cite{Qian2020}.
More recently, a reconfigurable DON based on optoelectronic fused computing architecture has been proposed\cite{Zhou2021}, which can perform different neural networks and achieve a high model complexity with millions of neurons. Although DONs have witnessed significant progress in the past few years, their functions mainly focus on image classification and object recognition without involving any interaction with the environment. To our knowledge, human-level AI based on DONs that can perform decision-making and control has not yet been developed.


In this work, we bring the capability of decision-making and control directly from high-dimensional sensory inputs to DON. The networks build upon deep reinforcement learning to interact with a simulated environment for optimal control policies. The training process of policy is based solely on deep reinforcement learning from self-play without dataset or guidance. A phase profile mapping features each layer of the DON and thus can be immediately implemented by optical modulation devices. The effectiveness of the proposed DON is validated with three typical games, Tic-Tac-Toe, Super Mario Bros., and Car Racing. We also provide a direct experimental demonstration of such DON capable of playing Tic-Tac-Toe. Excellent agreement can be found between theoretical prediction and experimental measurements. This work enables a fundamental shift from the target-driven control of a pre-designed state for simple recognition or classification tasks to human-imitative AI, revealing the potential of optoelectronic AI systems to solve complex real-world problems. We envision that such DONs find promising applications in autonomous driving, industrial robots, and intelligent manufacturing, aiming to enhance human life in every aspect.

\section{Results}\label{sec2}

\subsection{The network for decision-making and control}\label{subsec1}

\begin{figure*}[htbp]%
\centering
\includegraphics[width=0.9\textwidth]{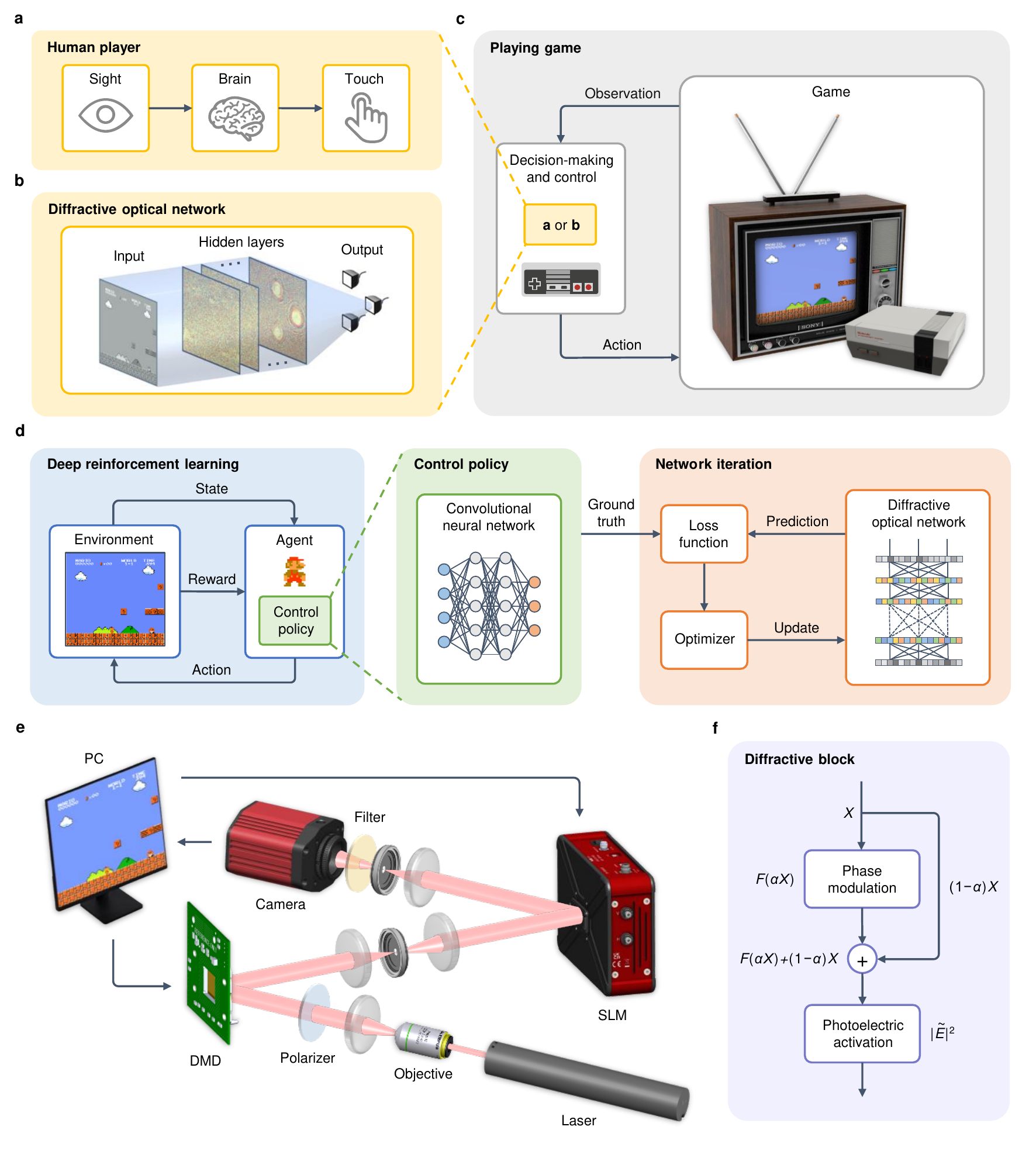}
\caption{\textbf{The DON for decision-making and control.} \textbf{a--c} The proposed network plays the video game of Super Mario Bros. in a human-like manner. In the network architecture, an input layer captures continuous, high-dimensional game snapshots (seeing), a series of diffractive layers choose a particular action through a learned control policy for each situation faced (making a decision), and an output layer maps the intensity distribution into preset action regions to generate the control signals in the games (controlling). \textbf{d} Training framework of policy and network. Deep reinforcement learning through an agent interacts with a simulated environment to find a near-optimal control policy represented by a CNN, which is employed as the ground truth to update the DON by error backpropagate algorithm. \textbf{e} The experimental setup of DON for decision-making and control. \textbf{f} The building block of DON.}\label{fig1}
\end{figure*}

The working principle of the DON for decision-making and control is illustrated in Fig. \ref{fig1}a--c, using an example of playing Nintendo’s classic video game Super Mario Bros. In general, a human player goes through seeing, understanding, and making a decision in each step, and these perception and control behaviors loop until the game is over. In order to play games in a human-like manner, the network necessitates the sensory capability to capture continuous, high-dimensional state spaces and the controllable execution ability of sequences of different behaviors. The DON shown in Fig. \ref{fig1}b comprises the specific free-space configuration: an input layer with images encoded using an optical modulation device, multiple hidden layers encoding phases of transmitted waves, and an output layer in which the computational results are imaged into.

More importantly, the proposed framework for decision-making and control integrates deep reinforcement learning and DON into a training procedure, allowing interaction between the game and the agent to learn control policies that can be implemented through the optical computing platform. The method observes each state within the game environment and chooses a particular action through a learned control policy for each situation. Then, the changed environment generates observation of the new state and makes the following action, and continuously updates the control policy in the loop. Unlike the previous optical networks, the input images from each video game frame are continuous high-dimensional sensory data. Furthermore, the execution procedure, such as playing games, is essentially a type of interactive control rather than the one-way recognition for a single objective, such as written digits or fashion items.

To address the complexity of imitating human players on the optical platform, we develop the training framework of policy and network shown in Fig. \ref{fig1}d, using a combination of novel and existing general-purpose techniques for neural network architectures. As shown in the middle block of Fig. \ref{fig1}d, central to the architecture is a control policy $\pi_\theta(a\vert s)$, which is represented by a convolutional neural network (CNN) with parameters $\theta$ that makes states $s$ as inputs and takes actions $a$ as outputs by optimizing the reward of games of self-play. Note that the training epoch of deep reinforcement learning is markedly more than that of the DONs due to the training of policies starting from entirely random behavior. Thus, we developed the training process approach with two main phases to eliminate unnecessary computations. Firstly, deep reinforcement learning through an agent interacts with a simulated game environment to find a near-optimal control policy to meet the specified goals. Secondly, the control policy updates the DON by the error backpropagation algorithm.

In the first phase, a deep reinforcement learning algorithm collects data to find a control policy concerning the specific reward function through interaction with the game environment, thereby achieving the desired outcome.
The states of these games need to satisfy the Markov property that the information of a particular state contains all relevant histories. Thus, it is possible to perform actions in the current state and move to the next state without considering the previous states. The agent interacts with the environment through a sequence of observations, actions, and rewards. At each step of interaction, the agent observes the state of the environment to decide on an action to take and then gives rewards based on the game result. The neural network decides the best action for each step based on the reward. It continuously updates the policy using proximal policy optimization (PPO)\cite{DBLP:journals/corr/SchulmanWDRK17} to find the optimal action. After testing, the trained policies can all complete the respective game. Compared with the previous studies, the algorithm only requires game rules without the need for human data, guidance, or domain knowledge, avoiding the performance's dependence on the dataset's quality.

In the second phase, the control policy is transferred onto the DON. The optimal control policy modeled using CNN is utilized as the ground truth during the learning procedure. Meanwhile, following the forward propagation model based on Huygens’ principle and Rayleigh-Sommerfeld diffraction, the encoded input light can be directed into any desired location at the output layer via the learnable transmission coefficients, that is, phase profiles of hidden layers in the network. The energy distributions clustered in the target detection region imply the prediction results. The transmission coefficients at each diffractive layer should be adequately trained via the error backpropagation algorithm and a loss function with mean square error (MSE), which is defined to evaluate the performance between the output intensities and the ground truth target. The adaptive moment estimation (Adam)\cite{DBLP:journals/corr/KingmaB14}, an algorithm for first-order gradient-based optimization of stochastic objective functions, is adopted to reduce the loss function. Then, the gradient of the loss function concerning all the trainable network variables is backpropagated to iteratively update the network during each cycle of the training phase until the network converges. 


Once the training is completed, the target phase profiles of the diffractive layers are determined, which are ready to connect the physical and digital worlds for optical neuromorphic computing. Here, we choose an approach similar to the diffractive processing unit\cite{Zhou2021} to build the network because of its reconfigurability and ability to support millions of neurons for computation. 
The experimental setup of the DON is shown in Fig. \ref{fig1}e. 
The entire computing process is primarily optical, except for the dataflow control. These light modulation devices are very fast and therefore allow for real-time computation.
Such an experimental system allows for a deep residual framework that can overcome the vanishing gradients problem by introducing shortcut connections between layers, and the architecture has become one of the cornerstones of neural networks\cite{7780459}. Fig. \ref{fig1}f demonstrates a block that composes the DON. First, when there is an angle between the polarization direction of incident light and the extraordinary axis of the liquid crystal of spatial light modulator (SLM), some light will not be modulated and reflected directly to the camera, thus creating a shortcut connection. Formally, the incident light is denoted as $X$, the diffraction computation is denoted as $F(X)$, and the original mapping can be recast into $F(\alpha X)+(1-\alpha)X$, where $\alpha$ is the modulation ratio of SLM, which can be fine-tuned by rotating laser and polarizer to change the polarization direction (or adding a half-wave plate). Compared to previous research\cite{Dou:20}, this approach does not require introducing additional optical devices, providing a free improvement. In addition, the approach also lowers the bar for the polarization state of light, and partially polarized light can be used in the network. Then, we use the photoelectric effect occurring at each image sensor pixel to implement the activation function of diffractive neurons, denoted as $\lvert\tilde{E}\rvert^2$. 
In addition, to some extent, the exposure of the camera and the differences in resolution between various devices can be analogized to the layer normalization and downsampling operations of neural networks, respectively. Unlike previous studies that used complex network structures, we stack the block to build the DON.

\subsection{Playing Tic-Tac-Toe}\label{subsec2}

\begin{figure*}[htbp]%
\centering
\includegraphics[width=0.9\textwidth]{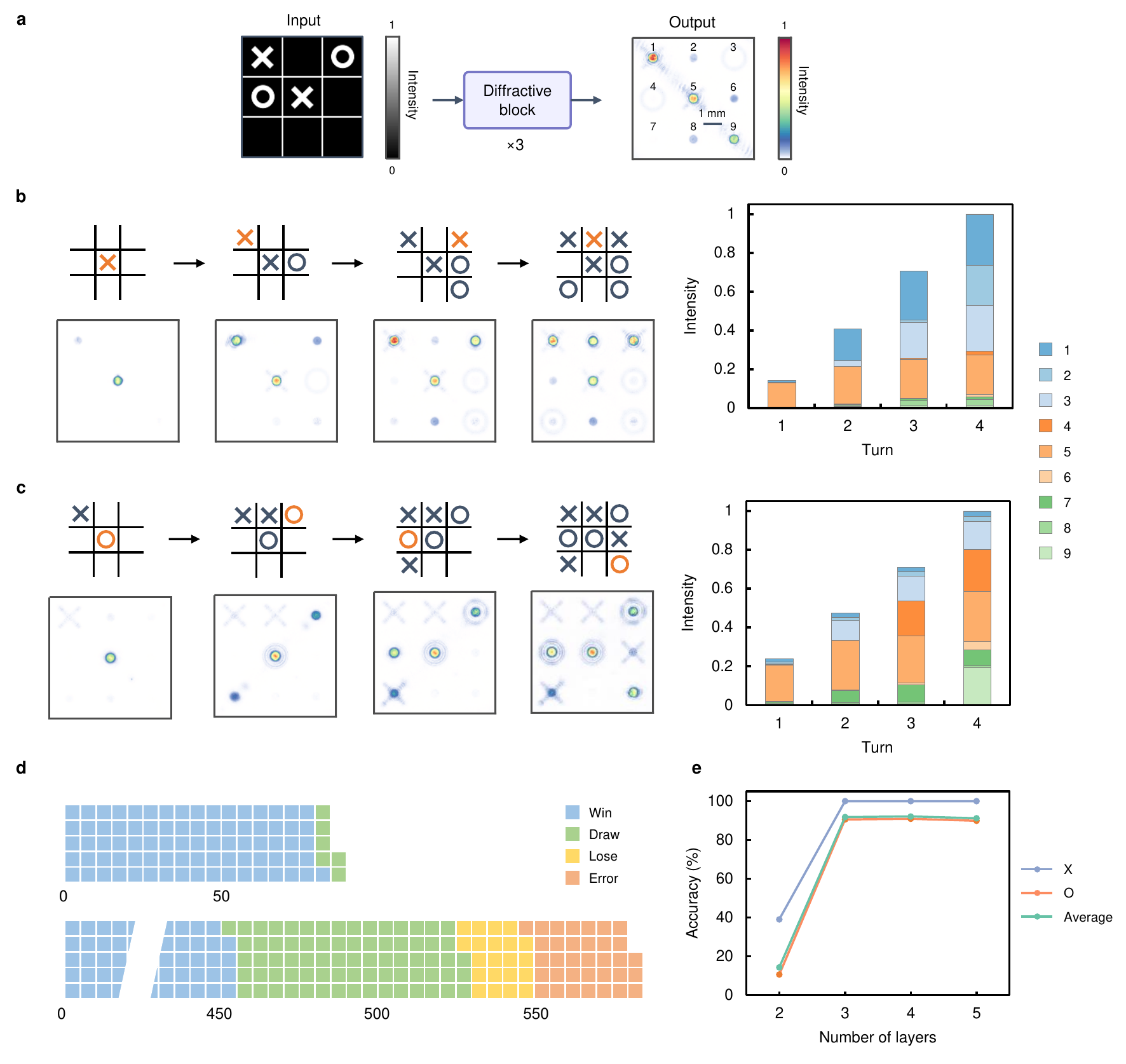}
\caption{\textbf{Playing Tic-Tac-Toe.} \textbf{a} The schematic illustration of the DON composed of an input layer, hidden layers of cascaded three diffractive blocks, and an output layer for playing Tic-Tac-Toe. \textbf{b,c} The sequential control of the DON in performing gameplay tasks for X and O, respectively. \textbf{d} The accuracy rate of playing Tic-Tac-Toe. There is a collection of 87 games utilized for predicting the X, obtaining 81 wins and 6 draws in these games. In the rest of the 583 games, the O obtains 454 wins, 74 draws, and 21 losses. When previous moves have occupied the predicted position at a turn, such a case is counted as a playing error and occurs 34 times. \textbf{e} Dependence of the prediction accuracy on the number of hidden layers.}\label{fig2}
\end{figure*}

In our first implementation, we perform the decision-making and control for Tic-Tac-Toe. This classic game is played on a 3$\times$3 grid of cells where each player places their mark, an X or an O, in an empty cell. The first player to place three of their marks in a row vertically, horizontally, or diagonally wins the game. If all cells are filled, and neither player has three marks in a row, the game is declared a draw. There are 255,168 possible ways to play this game, and we use the proposed network architecture to capture the effective policies to make the most optimal move in every possible situation.

To play this game, the network composed of three diffractive blocks is designed by the above training algorithm.
The input images carrying the information of the current states are encoded into the amplitude of the input field to the network. The network is trained to map the incident energy into nine cells corresponding to the grid (labeled by the number 1--9), where the received energy distribution at each region reveals the current state and predicts the probability of the player's next move, as shown in Fig. \ref{fig2}a. Since the observed state and the action are both discrete in this game, Tic-Tac-Toe can be considered to demonstrate our method for a collection of tasks with discrete state and action spaces.

Note that the first player (X) and the second player (O) have different control policies; specifically, X tries to win, and O tries to draw in the ideal case. 
After training, the X moves of each turn are illustrated in Fig. \ref{fig2}b. In the first turn, three possible positions, 1 and 5, are predicted as shown on the output. However, the starting position of 5 is finally chosen because of the maximum energy intensity among these positions. After the O responds to X, the input image changes, and in the second turn, the intensity distributions change as well so that the predicted move of X at the position of 1 is determined by extracting the maximum signal among the unoccupied positions. It is also noted that the output intensity is not only focused on the predicted position but also on the current states with occupied positions. Following this prediction and control procedure, the first player wins in the fourth turn. Following the same principle, the O moves are predicted and controlled, as shown in Fig. \ref{fig2}c. It can be observed that the O responds to a corner opening with a central mark and chooses moves next to the X to avoid the opponent having three marks in a row. In such a way, O avoids the win of X. This policy is successfully used in the proposed network, and a draw game result is shown in Fig. \ref{fig2}c, while O can win if X plays weakly in some exceptional cases.

While, in general, a human player aims to win the game, Tic-Tac-Toe will end in a draw if both players play their best because it is a zero-sum game. To evaluate the accuracy and effectiveness of our proposed network in playing Tic-Tac-Toe, we use the sum of the win and draw rate as the accuracy rate. After the self-play training per game rules, we numerically test the design of the DON with all possible states, as shown in Fig. \ref{fig2}d. The policy we trained is optimal and will only choose the best moves, so only 670 states will appear. Among them, the accuracy rate of X is 100$\%$, the accuracy rate of O is 90.56$\%$, and the average rate is 91.79$\%$. The accuracy of the network in predicting the O shows a slight degradation relative to that of X due to the factors such as more complex policy and more states of O.

In addition, we evaluate the dependence of the prediction accuracy on the number of hidden layers in Fig. \ref{fig2}e.
It can be seen that the accuracy of the network is greatly improved when changing from 2 layers to 3 layers because if there are not enough layers in the network, the shortcut connections between layers may not be fully computed, thus affecting the results.
However, the accuracy does not show a noticeable change when the layer number continues to increase from 3, which may be due to the following reasons. First, the DON is unsuitable for predicting states with high similarity\cite{Zheng:22}; see Supplementary Note 4 for detailed derivation. In addition, DONs have a similar global perceptual property to a multilayer perceptron (MLP), which can capture features at given spatial locations. However, it is difficult to capture features between different spatial locations\cite{NEURIPS2021_cba0a4ee}. We will discuss this point later in the paper.

\subsection{Playing Super Mario Bros.}\label{subsec3}

\begin{figure*}[htbp]%
\centering
\includegraphics[width=0.9\textwidth]{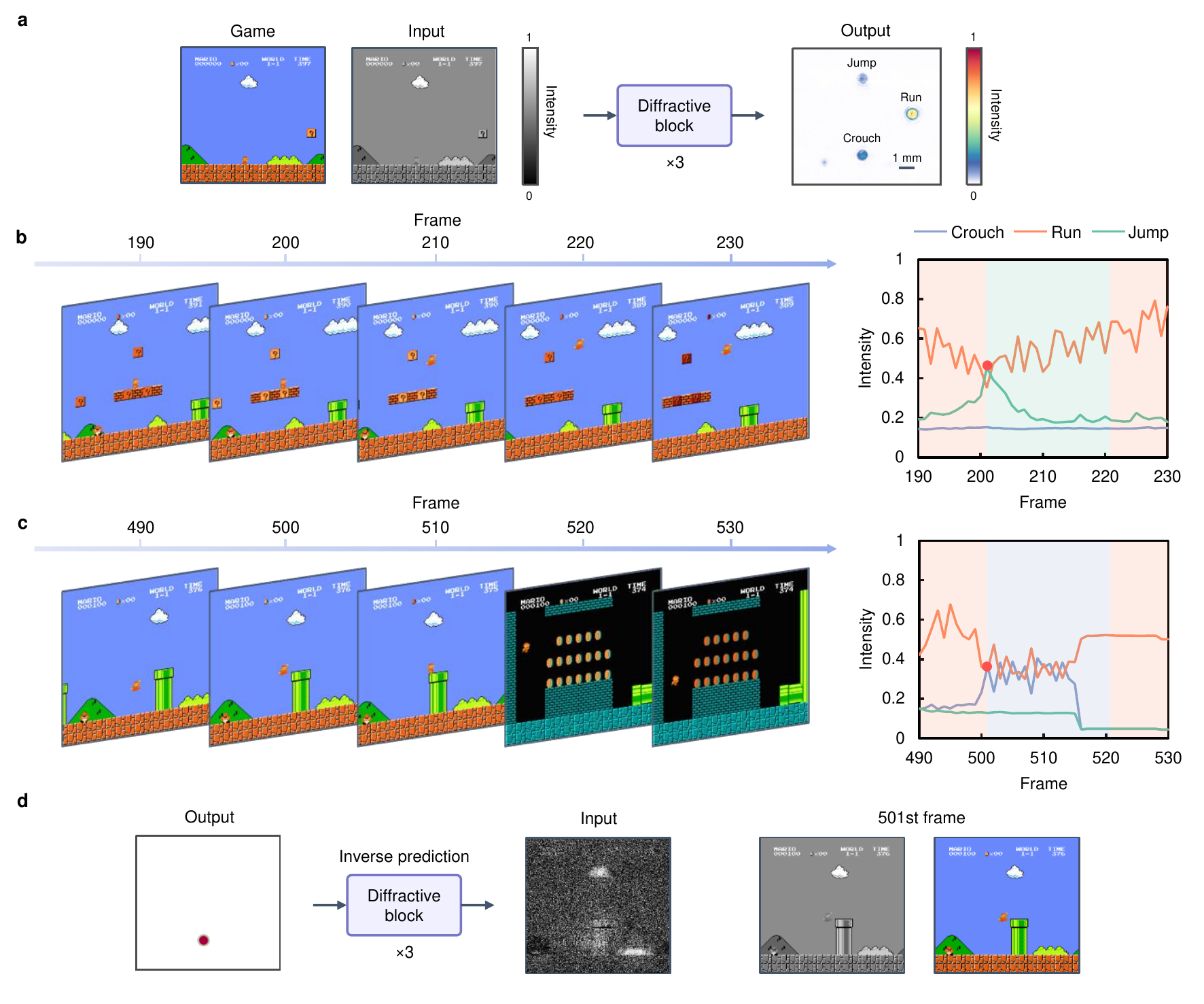}
\caption{\textbf{Playing Super Mario Bros.} \textbf{a} The layout of the designed network for playing Super Mario Bros. \textbf{b,c} Snapshots of Mario's jumping and crouching actions by comparing the output intensities of actions. The output intensity of the jump is maximum at the 201st frame, so the predicted action is jump, and Mario is controlled to act, shown in \textbf{b}. A similar series of prediction and control for another crouch action can also be observed in \textbf{c}. \textbf{d} The inverse prediction result. Considering the predicted crouch at the current state is crucial for updating Mario's action, we use the maximized output intensity of the crouch as input, ignoring the simultaneous output of other actions.}\label{fig3}
\end{figure*}

In our second implementation, the world 1-1 of the original Super Mario Bros. game is used to demonstrate the validity of DON. Unlike the Tic-Tac-Toe on a square-divided board, Super Mario Bros. is a video game with continuous high-dimensional state inputs. The gameplay consists of moving the player-controlled character, Mario, through two-dimensional levels to get to the level's end, traversing it from left to right, avoiding obstacles and enemies, and interacting with game objects. In the game, the player controls Mario to take discrete actions run, jump, and crouch. Under these considerations, this game can be an example of continuous state space and discrete action space for testing the proposed network.

Fig. \ref{fig3}a illustrates the DON for playing Super Mario Bros. The network consists of an input layer carrying the optical field encoded from each video game frame, hidden layers composed of cascaded three diffractive blocks trained by the same algorithm, and the output layer mapping the intensity distribution into preset regions. It is clear that the input images from the game scene consisting of moving backgrounds and different objects are more complex compared to the Tic-Tac-Toe with a regular pattern. In addition, the game images are similar between adjacent ones and constantly changing due to the gameplay on a side-scrolling platform, which challenges the DON in processing highly similar input states for choosing optimal actions.

After training with the control policy, the network makes decisions for Mario's optimal action. It achieves accurate control to reach the end of the level until taking down the flag raised above the castle, as shown in Supplementary Video 1. Specifically, at any given state, the most optimal action that Mario chooses to take is predicted by the maximum action signal. In the examples of Fig. \ref{fig3}b,c, we take some snapshots from Supplementary Video 1 to analyze the decision-making and control of Mario's actions in complex, time-varying configurations. Since the goal of our network is to finish the level as quickly as possible successfully, Mario should maintain the run action until the end while choosing to jump or crouch to overcome the challenges at certain states. Thus, the output intensity of run keeps high throughout the game, while the intensity of jump and crouch shows smaller fluctuations, verified by Fig. \ref{fig3}b,c. Although this significant intensity triggers the prediction only at a particular frame, this control signal is intentionally set to last for 20 frames to ensure Mario's finishing the entire action. It is worth noting that the intensity-frame curve remains relatively stable during the 516th to 530th frame, which can be understood with the static and high-contrast background images after Mario enters the pipe, as shown in Fig. \ref{fig3}c.

To gain an insight into how the DON makes decisions, we investigate the network's perception capability, employing inverse prediction in Fig. \ref{fig3}d. We demonstrate what the network has learned from the high-dimensional sensory input to perform the crouch action corresponding to the 501st frame image. We use the error backpropagation algorithm in a retrained network to inversely predict the input image at this moment, where $\alpha$ = 1 in the network to avoid the effect of the residual structure; see Supplementary Note 5 for detailed derivation. The inversely predicted image matches the original input image of the 501st frame, especially the background, such as clouds and grasses. When humans play the game, they may ignore these backgrounds and focus only on the critical parts, such as Mario, enemies, and pipes. The inverse prediction of the whole scene highlights the capability of the network to extract global features instead of local ones; the property is the same as MLP, further verifying the perception capability of the network in capturing the global features to make decisions.

\subsection{Playing Car Racing}\label{subsec4}

\begin{figure*}[htbp]%
\centering
\includegraphics[width=0.9\textwidth]{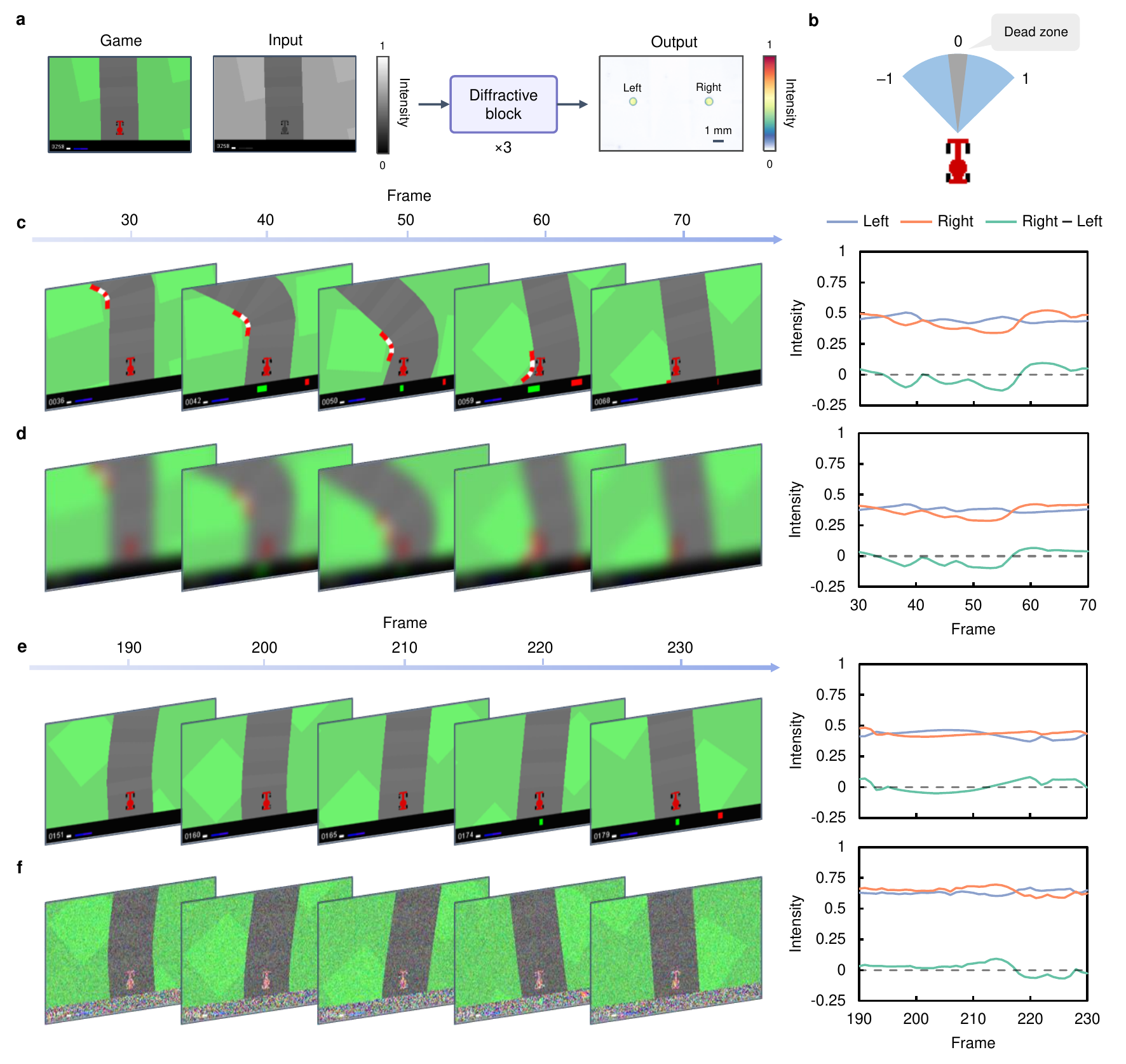}
\caption{\textbf{Playing Car Racing.} \textbf{a} The layout of the designed network for playing Car Racing. \textbf{b} The control of the steering direction and angle of the car with respect to the difference value between the intensities at the current state, normalized between $-$1 and 1. \textbf{c--f} Snapshots of controlling the car steering. When the car is facing a left-turn track in \textbf{c}, the output intensity on the left keeps the value greater than the right intensity, allowing continuous control in updating the rotation angle of the left-turn action. A similar control process can also be performed for the right-turn track in \textbf{e}. In addition, the anti-disturbance of the network is validated by introducing the Gaussian blur \textbf{d} and Gaussian noise \textbf{f} to the game images, respectively.}\label{fig4}
\end{figure*}

In our third implementation, we demonstrate the proposed network capability in Car Racing, which requires perceiving the game environment using continuous high-dimensional inputs and making decisions to control the car by performing continuous steering actions. The game’s control policy is trained based on the rules of keeping the car within the track by controlling its rotation, and the car is set to increase the speed once the game starts continuously. The DON architecture is shown in Fig. \ref{fig4}a similar to previous examples. The input energy of the optical field is redistributed through three diffractive blocks into the two designated regions on the left and right of the output layer. The difference value between the intensities at the current state controls the steering direction and angle of the car shown in Fig. \ref{fig4}b. In addition, just as in the steering dead zone in real vehicles, a slight difference value would not lead to steering action to avoid disturbance.

The successful network implementation in Car Racing is illustrated in Supplementary Video 2, where the car is controlled in the center of the track almost within the whole lap. For the two basic actions of the left and right turn, some exemplary snapshots are provided in Fig. \ref{fig4}c,e. Specifically, the negative difference values in Fig. \ref{fig4}c predict the left turn of the car wheel, while the larger absolute values indicate sharper turns. It is also observed that sometimes the difference values approach zero, and a rotation angle of 0 is predicted to keep the car moving in the direction of the current state. Due to the larger turning angle of the track, the intensity difference of the left turn shows a more drastic change. It is also intriguing that although the steering of the car in the left turn is somewhat unsmooth so that it does not appear in the middle of the track in certain states, the controlled action that is updated in the following state leads to successful gameplay. This real-time feedback and updating feature shows the great potential of the architecture for challenging auto-driving almost at the speed of light\cite{Rodrigues2021}, such as dealing with sudden obstacles. 

To validate the anti-disturbance ability of the proposed approach, we introduce two crucial randomization disturbance mechanisms to the frame image of the game and then test the network performance in controlling Car Racing. With the same network previously trained, Gaussian blur and Gaussian noise are respectively added to the frames, and the control results are shown in Fig. \ref{fig4}d and Fig. \ref{fig4}f, respectively. 
Although the introduction of disturbances, including blur and noise, causes the quality decline of the input image, the car can still maintain accurate and effective control to successfully complete the game, as verified by Supplementary Videos 3 and 4. Compared with the normal cases in Fig. \ref{fig4}c,e, the output intensity curves in Fig. \ref{fig4}d,f show similar trends to control the left or right turning actions. However, the curves are less smooth, with more amplitude fluctuations, indicating unsmooth steering angle control. The successful control in the cases with the randomization disturbance reveals the great perception of the game environment, especially the full access to the global features.

\subsection{Experimental demonstration of playing Tic-Tac-Toe}\label{subsec4}

\begin{figure*}[htbp]%
\centering
\includegraphics[width=0.9\textwidth]{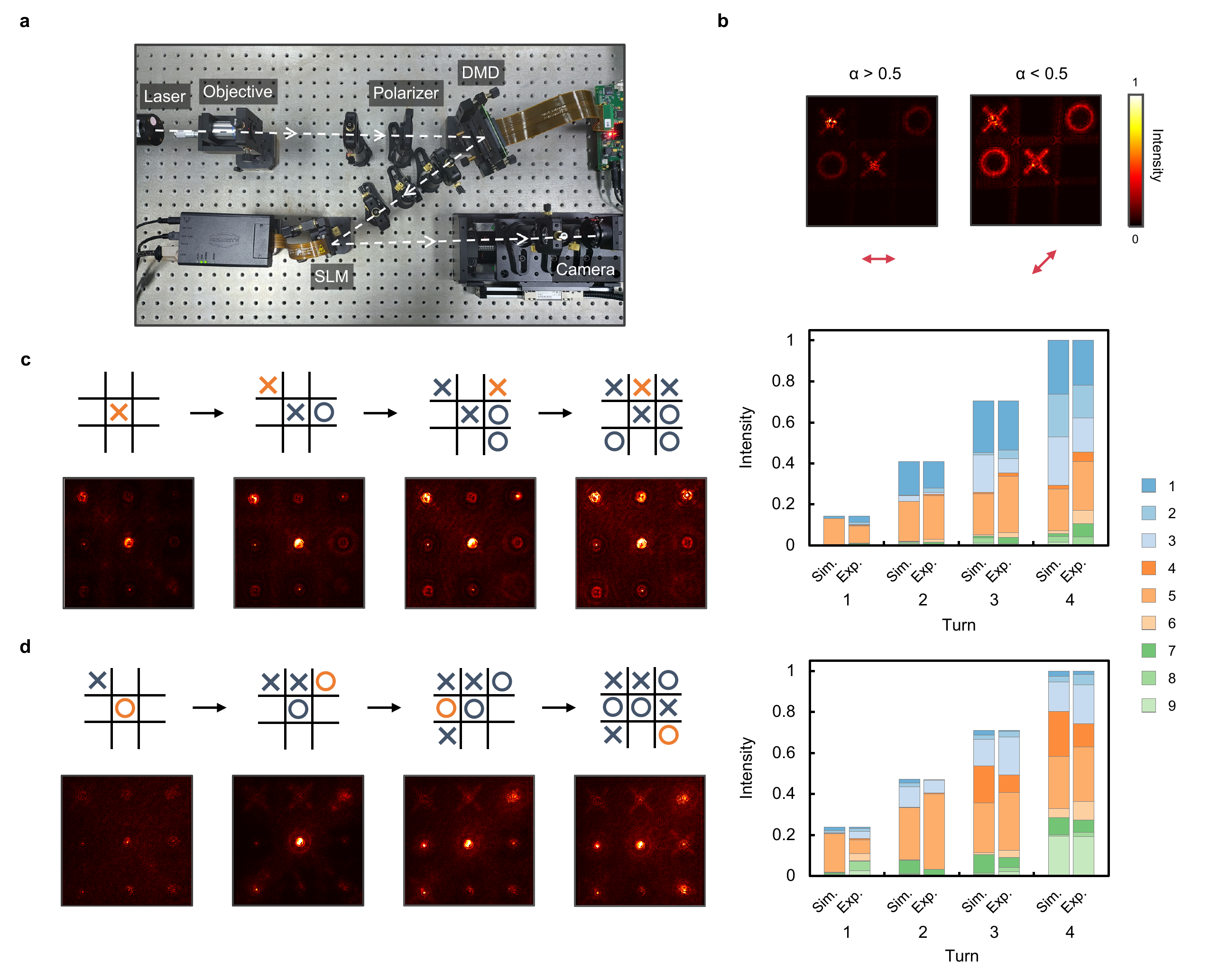}
\caption{\textbf{Experimental demonstration of the DON for Tic-Tac-Toe.} \textbf{a} The photo of the experimental system, where the unlabeled devices are lenses, a spatial filter is used to remove the unwanted multiple-order energy peaks, and a filter is mounted on the camera. \textbf{b} The output of the first layer of the sample in Fig. \ref{fig2}a, and the red arrows represent the polarization direction of incident light. \textbf{c,d} The sequential control of the DON in playing the same two games as in Fig. \ref{fig2}b,c, respectively. The experimental results are normalized based on simulation results. Sim. simulation, Exp. experimental.}\label{fig5}
\end{figure*}

Finally, to test the real experimental performance of the DON, we built an experimental system using off-the-shelf optical modulation devices. We tested it by playing Tic-Tac-Toe, as shown in Fig. \ref{fig5}a. A laser beam with a working wavelength of 632.8nm is expanded using a microscope objective and lens, while a linear polarizer can be embedded to adjust the incident light intensity, which is then projected onto a digital micromirror device (DMD). The input image data is optically encoded and modulated by the DMD, followed by two relay lenses to adjust the image to the appropriate size and projected onto the SLM for phase modulation. The optical iris is used to filter out high-order diffractions and stray light. The diffraction pattern is imaged onto the camera; then, the output image is input to the DMD for the next diffractive layer until the end of the network computation. After that, the optical intensities in the predefined detection zones are extracted from the output image, and the predicted results are decoded to generate the control signals in the games. Then the new frame image of the video game stimulates the new process procedure, and the updated results control the game until the end. In addition, because DMD is a binary device, the training process needs to simulate the fast rotation of the micromirror when displaying grayscale images to make the training results more practical. We adapt the previously trained phase profiles for DMD, as detailed in Supplementary Note 6. 

We first test our proposed residual architecture, Fig. \ref{fig5}b shows the effect of our proposed residual architecture, which is the output of the first layer of the sample in Fig. \ref{fig2}a. It can be seen that the value of $\alpha$ varies with the polarization direction of the incident light. This shows that our proposed residual architecture is valid and can be flexibly adapted to the task.

After that, we tested the same two games as in Fig. \ref{fig2}b,c, and the experimental results are presented in \ref{fig5}c,d. It can be seen that the intensity distribution of the output changes as the input game state changes. Due to the unavoidable physical error in the experimental system, the experimental results are different from the simulated ones, but the overall intensity changes are very similar. The maximum intensity distributions occur at the same positions, and the same games are successfully completed.

\section{Discussion}\label{sec3}

We have demonstrated DONs for decision-making and control. The optimal control policy enables this technique through a harmonious combination of deep reinforcement learning and the DON architecture. Based solely on reinforcement learning from self-play, the control policy of the training algorithm is flexible, as demonstrated by successfully learning to play the three types of classic games. In addition, we further exploit the potential of the photoelectric fusion DON by introducing a free residual architecture, which achieves excellent performance in the simplest network structure.

It is worth noting that Tic-Tac-Toe does not achieve perfect results despite the definite rules and optimal control policy, just like Super Mario Bros. and Car Racing. There are several possible reasons for this result:
Playing Tic-Tac-Toe needs to strategically handle different states and a more significant number of output signals.
The gameplay of Tic-Tac-Toe requires correct predictions at each state, while the other two games show better error tolerance and accidental mistakes do not necessarily affect the results. In addition, using the difference as a mechanism to trigger actions also improves the network's performance in Car Racing to some extent.
Since the DON is not good at extracting local features, the differences in intensity distributions between the adjacent input board images are challenging to detect for Tic-Tac-Toe. 



By testing our proposed DON on the challenging domain of classic games, we demonstrate its ability to master difficult game control policies for playing game, which is also the first time on an optical platform.
This work bridges the gap between optical and digital neural networks aiming to achieve human-level AI. The most important aspect is that the decision-making and control process is implemented in optical devices at the speed of light by imitating human competence. 
Another ideal platform for implementing DONs is metasurface. Metasurfaces provide an unprecedented ability to manipulate the wavefront of light and are widely used to implement sophisticated functions such as holography and computational imaging\cite{doi:10.1126/science.abi6860,Li2019,PhysRevApplied.18.044078,Padilla2022}.
Therefore, driven by the demand of all-optical on-chip integration of AI systems, some recent studies have introduced optical metasurfaces consisting of an array of subwavelength meta-atoms to replace bulky diffractive optical devices for high-density integration\cite{Liu2022,Luo2022,https://doi.org/10.1002/advs.202204699,Qian2022}.
The working mechanism and design principle of our proposed DONs are universal and thus can be generalized to nanostructures. We have also implemented the above network on metasurfaces; see Supplementary Note 7 for details. Therefore, a metasurface-based DON can be envisaged and will serve as a very promising candidate for photonic integrated circuits.



Despite the exciting results of playing games, the DON currently has limitations for handling more complex tasks. First, for the sake of the computational requirements of optical forward propagation, we deploy a two-phase training architecture to obtain the policy model before iterating the DON instead of end-to-end learning in this work. Combining the two steps may reduce errors and make it easier to use. 
Second, ideally, the last layer of the network should not have shortcut connection, which can be improved by modifying the experimental system.
In addition, given the similar properties of DON and MLP, the introduction of MLP-based attention mechanisms\cite{NEURIPS2021_cba0a4ee,9912362} into the field of optics could be considered. 
Moreover, the inference and control capability of DONs could also be improved by introducing methods such as nonlinear optical effects\cite{doi:10.1021/acs.nanolett.0c01105,Xiao2018,8350283,Zuo:19}, multichannel structures\cite{Xu2022}, and Fourier space\cite{PhysRevLett.123.023901} in the future, leading to a variety of new applications. While preliminary, this research suggests that the DON has great potential for processing complex visual inputs and tasks. It could provide a promising avenue for an optical computing system for decision-making and control, which would be a fruitful area for next-generation AI.

\section{Methods}\label{sec4}
\subsection{Experimental setup}\label{subsec7}
A He-Ne laser (HNL100L, Thorlabs) is used as the light source. A DMD (DLP7000, Texas Instruments) mounted on a controller board (F4100, X-digit), consists of 1,024$\times$768 micromirrors with a pitch of 13.68 $\mu$m, which can modulate the incident light at a maximum speed of 18 kHz and display 8-bit grayscale images. A liquid-crystal-on-silicon (LCoS) SLM (PLUTO-2.1, HOLOEYE Photonics) with a pixel size of 8 $\mu$m and a resolution of 1,920$\times$1,080 is used to modulate the phase of the wavefront. A grayscale CMOS camera (GS3-U3-51S5M, Teledyne FLIR) with a resolution of 2,448$\times$2,048 and a pixel size of 3.45 $\mu$m is used to capture the output. Self-developed Python scripts control the above devices.

\subsection{Training of network}\label{subsec7}
We use the open-source machine learning framework Pytorch to build the training algorithm. The training process is implemented using Python (v3.9.13), Gym (v0.21.0) environment, and Pytorch (v1.11.0) framework on a desktop computer with Intel Core i7-13700K CPU, Nvidia GeForce RTX 4070 Ti GPU, and 16 GB RAM.

We train each DON individually. The structure of each network is almost identical, except that the number of neurons in each layer is determined by the resolution of the SLM and the aspect ratio of the game, which are 1,080$\times$1,080, 1,152$\times$1,080, and 1,620$\times$1,080 for Tic-Tac-Toe, Super Mario Bros., and Car Racing respectively. In the iterative process of the DON, $\alpha$ is 0.5, the learning rate is 0.01, the number of training epochs is 1,000, and constrain the phase-modulation range to 0--2$\pi$ for training. In addition, due to the unbalanced number of occurrences of output action, such as jump and crouch are far rarer than running in Super Mario Bros., we balance these actions by altering the weight each training example carries when computing the loss. Training of the network takes about tens of minutes. After training, the performance of networks is verified by interacting with the game environment. The phase profiles of networks are shown in the Supplementary Fig. 9--11.

In addition, parameter initialization, as the starting point of network training, determines the initial training position. The choice greatly affects the convergence speed and the final training results. We evaluate the effect of different parameter initializations on the results. The results show that the normal initialization works best; see Supplementary Fig. 4 for details.

\subsection{Modeling of control policy}\label{subsec8}

We train different control policies on each game using the same network architecture, learning algorithms, and hyperparameter settings. That shows our approach is robust enough to work on various games while incorporating only minimal prior knowledge. PPO uses two neural network architectures to design and optimize the policy: the critic network and the actor-network. Both networks are adapted during training, but only the actor-network is deployed as the control policy. Specifically, for the actor-network, The input to the CNN consists of a 84$\times$84 grayscale image produced by the preprocessing. The first hidden layer convolves 32 filters of 8$\times$8 with stride 4 with the input image. The second hidden layer convolves 64 filters of 4$\times$4 with stride 2. This is followed by a third convolutional layer convolving 64 filters of 3$\times$3 with stride 1. The final hidden layer is fully-connected with 512 latents. Each hidden layer is followed by a rectified linear unit (ReLU). The output layer is a fully-connected linear layer with a single output for each valid action. The critic network is also a CNN which is about the same as the actor-network, except the output represents the discounted expected future reward for various actions. The reward functions are detailed in Supplementary Note 3. The learning rate is 0.001, and the discount is 0.99. In addition, because consecutive frames do not vary much, we can skip 4 intermediate frames without losing much information to accelerate training. Train of the control policy takes about an hour to a dozen hours. After the training, we save each game frame and the corresponding action for DON training.




\section*{Data availability}

All data are available in the main text or the supplementary materials.

\section*{Code availability}

The codes used in this work is available in the GitHub repository at \url{https://github.com/qiujumin/DON-RL}



\begin{thebibliography}{10}
\expandafter\ifx\csname url\endcsname\relax
  \def\url#1{\burl{#1}}\fi
\expandafter\ifx\csname urlprefix\endcsname\relax\def\urlprefix{URL }\fi
\providecommand{\bibinfo}[2]{#2}
\providecommand{\eprint}[2][]{\url{#2}}
\providecommand{\doi}[1]{\url{https://doi.org/#1}}
\bibcommenthead

\bibitem{10.1145/3065386}
\bibinfo{author}{Krizhevsky, A.}, \bibinfo{author}{Sutskever, I.} \&
  \bibinfo{author}{Hinton, G.~E.}
\newblock \bibinfo{title}{Imagenet classification with deep convolutional
  neural networks}.
\newblock \emph{\bibinfo{journal}{Commun. ACM}}
  \textbf{\bibinfo{volume}{60}}~(6), \bibinfo{pages}{84–90}
  (\bibinfo{year}{2017}).

\bibitem{Russakovsky2015}
\bibinfo{author}{Russakovsky, O.} \emph{et~al.}
\newblock \bibinfo{title}{Imagenet large scale visual recognition challenge}.
\newblock \emph{\bibinfo{journal}{International Journal of Computer Vision}}
  \textbf{\bibinfo{volume}{115}}~(3), \bibinfo{pages}{211--252}
  (\bibinfo{year}{2015}).

\bibitem{chen-etal-2017-enhanced}
\bibinfo{author}{Chen, Q.} \emph{et~al.}
\newblock \bibinfo{title}{Enhanced {LSTM} for natural language inference}, In \emph{\bibinfo{booktitle}{Proceedings of the 55th Annual Meeting of the Association for Computational Linguistics}}, \bibinfo{pages}{1657--1668} (\bibinfo{year}{2017}).

\bibitem{devlin-etal-2019-bert}
\bibinfo{author}{Devlin, J.}, \bibinfo{author}{Chang, M.-W.},
  \bibinfo{author}{Lee, K.} \& \bibinfo{author}{Toutanova, K.}
\newblock \bibinfo{title}{{BERT}: Pre-training of deep bidirectional transformers for language understanding}, In \emph{\bibinfo{booktitle}{2019 Annual Conference of the North American Chapter of the Association for Computational Linguistics}}, \bibinfo{pages}{4171--4186} (\bibinfo{year}{2019}).

\bibitem{10.1145/2939672.2939754}
\bibinfo{author}{Grover, A.} \& \bibinfo{author}{Leskovec, J.}
\newblock \bibinfo{title}{Node2vec: Scalable feature learning for networks}, In \emph{\bibinfo{booktitle}{Proceedings of the 22nd ACM SIGKDD International Conference on Knowledge Discovery and Data Mining}}, \bibinfo{pages}{855–864} (\bibinfo{year}{2016}).

\bibitem{Ma2021}
\bibinfo{author}{Ma, W.} \emph{et~al.}
\newblock \bibinfo{title}{Deep learning for the design of photonic structures}.
\newblock \emph{\bibinfo{journal}{Nature Photonics}}
  \textbf{\bibinfo{volume}{15}}~(2), \bibinfo{pages}{77--90}
  (\bibinfo{year}{2021}).

\bibitem{https://doi.org/10.1002/adom.202200097}
\bibinfo{author}{Khatib, O.}, \bibinfo{author}{Ren, S.},
  \bibinfo{author}{Malof, J.} \& \bibinfo{author}{Padilla, W.~J.}
\newblock \bibinfo{title}{Learning the physics of all-dielectric metamaterials
  with deep lorentz neural networks}.
\newblock \emph{\bibinfo{journal}{Advanced Optical Materials}}
  \textbf{\bibinfo{volume}{10}}~(13), \bibinfo{pages}{2200097}
  (\bibinfo{year}{2022}).

\bibitem{https://doi.org/10.1002/adfm.202101748}
\bibinfo{author}{Khatib, O.}, \bibinfo{author}{Ren, S.},
  \bibinfo{author}{Malof, J.} \& \bibinfo{author}{Padilla, W.~J.}
\newblock \bibinfo{title}{Deep learning the electromagnetic properties of
  metamaterials—a comprehensive review}.
\newblock \emph{\bibinfo{journal}{Advanced Functional Materials}}
  \textbf{\bibinfo{volume}{31}}~(31), \bibinfo{pages}{2101748}
  (\bibinfo{year}{2021}).

\bibitem{Huang20}
\bibinfo{author}{Huang, L.}, \bibinfo{author}{Xu, L.} \&
  \bibinfo{author}{Miroshnichenko, A.~E.}
\newblock \bibinfo{title}{\textit{Deep learning enabled nanophotonics}}.
  in \emph{\bibinfo{booktitle}{Advances and Applications in Deep Learning}} (ed.\bibinfo{editor}{Aceves-Fernandez, M.~A.})
  Ch.~\bibinfo{chapter}{4} (\bibinfo{publisher}{IntechOpen},
  \bibinfo{address}{London}, \bibinfo{year}{2020}).

\bibitem{Nadell:19}
\bibinfo{author}{Nadell, C.~C.}, \bibinfo{author}{Huang, B.},
  \bibinfo{author}{Malof, J.~M.} \& \bibinfo{author}{Padilla, W.~J.}
\newblock \bibinfo{title}{Deep learning for accelerated all-dielectric
  metasurface design}.
\newblock \emph{\bibinfo{journal}{Opt. Express}}
  \textbf{\bibinfo{volume}{27}}~(20), \bibinfo{pages}{27523--27535}
  (\bibinfo{year}{2019}).

\bibitem{10.1117/1.AP.2.2.026003}
\bibinfo{author}{Xu, L.} \emph{et~al.}
\newblock \bibinfo{title}{{Enhanced light–matter interactions in dielectric
  nanostructures via machine-learning approach}}.
\newblock \emph{\bibinfo{journal}{Advanced Photonics}}
  \textbf{\bibinfo{volume}{2}}~(2), \bibinfo{pages}{026003}
  (\bibinfo{year}{2020}).

\bibitem{doi:10.1021/acs.nanolett.9b03971}
\bibinfo{author}{Wiecha, P.~R.} \& \bibinfo{author}{Muskens, O.~L.}
\newblock \bibinfo{title}{Deep learning meets nanophotonics: A generalized
  accurate predictor for near fields and far fields of arbitrary 3d
  nanostructures}.
\newblock \emph{\bibinfo{journal}{Nano Letters}}
  \textbf{\bibinfo{volume}{20}}~(1), \bibinfo{pages}{329--338}
  (\bibinfo{year}{2020}).

\bibitem{Dai:21}
\bibinfo{author}{Dai, P.} \emph{et~al.}
\newblock \bibinfo{title}{Accurate inverse design of Fabry-Perot-cavity-based
  color filters far beyond sRGB via a bidirectional artificial neural network}.
\newblock \emph{\bibinfo{journal}{Photon. Res.}}
  \textbf{\bibinfo{volume}{9}}~(5), \bibinfo{pages}{B236--B246}
  (\bibinfo{year}{2021}).






\bibitem{Shen2017}
\bibinfo{author}{Shen, Y.} \emph{et~al.}
\newblock \bibinfo{title}{Deep learning with coherent nanophotonic circuits}.
\newblock \emph{\bibinfo{journal}{Nature Photonics}}
  \textbf{\bibinfo{volume}{11}}~(7), \bibinfo{pages}{441--446}
  (\bibinfo{year}{2017}).

\bibitem{Feldmann2019}
\bibinfo{author}{Feldmann, J.}, \bibinfo{author}{Youngblood, N.},
  \bibinfo{author}{Wright, C.~D.}, \bibinfo{author}{Bhaskaran, H.} \&
  \bibinfo{author}{Pernice, W. H.~P.}
\newblock \bibinfo{title}{All-optical spiking neurosynaptic networks with
  self-learning capabilities}.
\newblock \emph{\bibinfo{journal}{Nature}}
  \textbf{\bibinfo{volume}{569}}~(7755), \bibinfo{pages}{208--214}
  (\bibinfo{year}{2019}).

\bibitem{PhysRevX.9.021032}
\bibinfo{author}{Hamerly, R.}, \bibinfo{author}{Bernstein, L.},
  \bibinfo{author}{Sludds, A.}, \bibinfo{author}{Solja\ifmmode \check{c}\else
  \v{c}\fi{}i\ifmmode~\acute{c}\else \'{c}\fi{}, M.} \&
  \bibinfo{author}{Englund, D.}
\newblock \bibinfo{title}{Large-scale optical neural networks based on
  photoelectric multiplication}.
\newblock \emph{\bibinfo{journal}{Phys. Rev. X}} \textbf{\bibinfo{volume}{9}},
  \bibinfo{pages}{021032} (\bibinfo{year}{2019}).

\bibitem{Zhang2021}
\bibinfo{author}{Zhang, H.} \emph{et~al.}
\newblock \bibinfo{title}{An optical neural chip for implementing
  complex-valued neural network}.
\newblock \emph{\bibinfo{journal}{Nature Communications}}
  \textbf{\bibinfo{volume}{12}}~(1), \bibinfo{pages}{457}
  (\bibinfo{year}{2021}).

\bibitem{Liu2021}
\bibinfo{author}{Liu, J.} \emph{et~al.}
\newblock \bibinfo{title}{Research progress in optical neural networks: theory,
  applications and developments}.
\newblock \emph{\bibinfo{journal}{PhotoniX}} \textbf{\bibinfo{volume}{2}}~(1),
  \bibinfo{pages}{5} (\bibinfo{year}{2021}).

\bibitem{Wu:20}
\bibinfo{author}{Wu, Z.}, \bibinfo{author}{Zhou, M.}, \bibinfo{author}{Khoram,
  E.}, \bibinfo{author}{Liu, B.} \& \bibinfo{author}{Yu, Z.}
\newblock \bibinfo{title}{Neuromorphic metasurface}.
\newblock \emph{\bibinfo{journal}{Photon. Res.}}
  \textbf{\bibinfo{volume}{8}}~(1), \bibinfo{pages}{46--50}
  (\bibinfo{year}{2020}).


\bibitem{doi:10.1126/science.aat8084}
\bibinfo{author}{Lin, X.} \emph{et~al.}
\newblock \bibinfo{title}{All-optical machine learning using diffractive deep
  neural networks}.
\newblock \emph{\bibinfo{journal}{Science}}
  \textbf{\bibinfo{volume}{361}}~(6406), \bibinfo{pages}{1004--1008}
  (\bibinfo{year}{2018}).

\bibitem{CHEN20211483}
\bibinfo{author}{Chen, H.} \emph{et~al.}
\newblock \bibinfo{title}{Diffractive deep neural networks at visible
  wavelengths}.
\newblock \emph{\bibinfo{journal}{Engineering}}
  \textbf{\bibinfo{volume}{7}}~(10), \bibinfo{pages}{1483--1491}
  (\bibinfo{year}{2021}).

\bibitem{Luo2022}
\bibinfo{author}{Luo, X.} \emph{et~al.}
\newblock \bibinfo{title}{Metasurface-enabled on-chip multiplexed diffractive
  neural networks in the visible}.
\newblock \emph{\bibinfo{journal}{Light: Science {\&} Applications}}
  \textbf{\bibinfo{volume}{11}}~(1), \bibinfo{pages}{158}
  (\bibinfo{year}{2022}).

\bibitem{Liu2022}
\bibinfo{author}{Liu, C.} \emph{et~al.}
\newblock \bibinfo{title}{A programmable diffractive deep neural network based
  on a digital-coding metasurface array}.
\newblock \emph{\bibinfo{journal}{Nature Electronics}}
  \textbf{\bibinfo{volume}{5}}~(2), \bibinfo{pages}{113--122}
  (\bibinfo{year}{2022}).

\bibitem{doi:10.1126/sciadv.abo6410}
\bibinfo{author}{Zheng, H.} \emph{et~al.}
\newblock \bibinfo{title}{Meta-optic accelerators for object classifiers}.
\newblock \emph{\bibinfo{journal}{Science Advances}}
  \textbf{\bibinfo{volume}{8}}~(30), \bibinfo{pages}{eabo6410}
  (\bibinfo{year}{2022}).

\bibitem{PhysRevLett.123.023901}
\bibinfo{author}{Yan, T.} \emph{et~al.}
\newblock \bibinfo{title}{Fourier-space diffractive deep neural network}.
\newblock \emph{\bibinfo{journal}{Phys. Rev. Lett.}}
  \textbf{\bibinfo{volume}{123}}, \bibinfo{pages}{023901}
  (\bibinfo{year}{2019}).

\bibitem{Qian2020}
\bibinfo{author}{Qian, C.} \emph{et~al.}
\newblock \bibinfo{title}{Performing optical logic operations by a diffractive
  neural network}.
\newblock \emph{\bibinfo{journal}{Light: Science {\&} Applications}}
  \textbf{\bibinfo{volume}{9}}~(1), \bibinfo{pages}{59} (\bibinfo{year}{2020}).

\bibitem{Zhou2021}
\bibinfo{author}{Zhou, T.} \emph{et~al.}
\newblock \bibinfo{title}{Large-scale neuromorphic optoelectronic computing
  with a reconfigurable diffractive processing unit}.
\newblock \emph{\bibinfo{journal}{Nature Photonics}}
  \textbf{\bibinfo{volume}{15}}~(5), \bibinfo{pages}{367--373}
  (\bibinfo{year}{2021}).

\bibitem{Xu2022}
\bibinfo{author}{Xu, Z.}, \bibinfo{author}{Yuan, X.}, \bibinfo{author}{Zhou,
  T.} \& \bibinfo{author}{Fang, L.}
\newblock \bibinfo{title}{A multichannel optical computing architecture for
  advanced machine vision}.
\newblock \emph{\bibinfo{journal}{Light: Science {\&} Applications}}
  \textbf{\bibinfo{volume}{11}}~(1), \bibinfo{pages}{255}
  (\bibinfo{year}{2022}).

\bibitem{DBLP:journals/corr/SchulmanWDRK17}
\bibinfo{author}{Schulman, J.}, \bibinfo{author}{Wolski, F.},
  \bibinfo{author}{Dhariwal, P.}, \bibinfo{author}{Radford, A.} \&
  \bibinfo{author}{Klimov, O.}
\newblock \bibinfo{title}{Proximal policy optimization algorithms}.
\newblock
  \bibinfo{eprint}{Preprint at {\href{https://arxiv.org/abs/1707.06347}{{1707.06347}}}} (\bibinfo{year}{2017}).

\bibitem{DBLP:journals/corr/KingmaB14}
\bibinfo{author}{Kingma, D.~P.} \& \bibinfo{author}{Ba, J.}
  \bibinfo{title}{Adam: {A} method for stochastic optimization}. In \emph{\bibinfo{booktitle}{3rd International Conference on Learning Representations}} (\bibinfo{year}{2015}).

\bibitem{7780459}
\bibinfo{author}{He, K.}, \bibinfo{author}{Zhang, X.}, \bibinfo{author}{Ren,
  S.} \& \bibinfo{author}{Sun, J.}
\newblock \emph{\bibinfo{title}{Deep residual learning for image recognition}}, In \emph{\bibinfo{booktitle}{2016 IEEE Conference on Computer Vision and Pattern Recognition}}, \bibinfo{pages}{770--778} (\bibinfo{year}{2016}).

\bibitem{Dou:20}
\bibinfo{author}{Dou, H.} \emph{et~al.}
\newblock \bibinfo{title}{Residual D2NN: training diffractive deep neural
  networks via learnable light shortcuts}.
\newblock \emph{\bibinfo{journal}{Opt. Lett.}}
  \textbf{\bibinfo{volume}{45}}~(10), \bibinfo{pages}{2688--2691}
  (\bibinfo{year}{2020}).

\bibitem{Zheng:22}
\bibinfo{author}{Zheng, S.}, \bibinfo{author}{Xu, S.} \& \bibinfo{author}{Fan,
  D.}
\newblock \bibinfo{title}{Orthogonality of diffractive deep neural network}.
\newblock \emph{\bibinfo{journal}{Opt. Lett.}}
  \textbf{\bibinfo{volume}{47}}~(7), \bibinfo{pages}{1798--1801}
  (\bibinfo{year}{2022}).

\bibitem{NEURIPS2021_cba0a4ee}
\bibinfo{author}{Tolstikhin, I.~O.} \emph{et~al.}
\emph{\bibinfo{title}{Mlp-mixer: An all-mlp architecture for vision}}, In \emph{\bibinfo{booktitle}{Advances in Neural Information Processing Systems}}, \textbf{\bibinfo{volume}{34}},
\bibinfo{pages}{24261--24272}
  (\bibinfo{year}{2021}).

\bibitem{Rodrigues2021}
\bibinfo{author}{Rodrigues, S.~P.} \emph{et~al.}
\newblock \bibinfo{title}{Weighing in on photonic-based machine learning for
  automotive mobility}.
\newblock \emph{\bibinfo{journal}{Nature Photonics}}
  \textbf{\bibinfo{volume}{15}}~(2), \bibinfo{pages}{66--67}
  (\bibinfo{year}{2021}).

\bibitem{doi:10.1126/science.abi6860}
\bibinfo{author}{Dorrah, A.~H.} \& \bibinfo{author}{Capasso, F.}
\newblock \bibinfo{title}{Tunable structured light with flat optics}.
\newblock \emph{\bibinfo{journal}{Science}}
  \textbf{\bibinfo{volume}{376}}~(6591), \bibinfo{pages}{eabi6860}
  (\bibinfo{year}{2022}).





\bibitem{Li2019}
\bibinfo{author}{Li, L.} \emph{et~al.}
\newblock \bibinfo{title}{Machine-learning reprogrammable metasurface imager}.
\newblock \emph{\bibinfo{journal}{Nature Communications}}
  \textbf{\bibinfo{volume}{10}}~(1), \bibinfo{pages}{1082}
  (\bibinfo{year}{2019}).


\bibitem{PhysRevApplied.18.044078}
\bibinfo{author}{Liu, T.}, \bibinfo{author}{Han, Z.}, \bibinfo{author}{Duan,
  J.} \& \bibinfo{author}{Xiao, S.}
\newblock \bibinfo{title}{Phase-change metasurfaces for dynamic image display
  and information encryption}.
\newblock \emph{\bibinfo{journal}{Phys. Rev. Applied}}
  \textbf{\bibinfo{volume}{18}}, \bibinfo{pages}{044078}
  (\bibinfo{year}{2022}).

\bibitem{Padilla2022}
\bibinfo{author}{Padilla, W.~J.} \& \bibinfo{author}{Averitt, R.~D.}
\newblock \bibinfo{title}{Imaging with metamaterials}.
\newblock \emph{\bibinfo{journal}{Nature Reviews Physics}}
  \textbf{\bibinfo{volume}{4}}~(2), \bibinfo{pages}{85--100}
  (\bibinfo{year}{2022}).





\bibitem{https://doi.org/10.1002/advs.202204699}
\bibinfo{author}{Wang, Z.}, \bibinfo{author}{Qian, C.}, \bibinfo{author}{Fan,
  Z.} \& \bibinfo{author}{Chen, H.}
\newblock \bibinfo{title}{Arbitrary polarization readout with dual-channel
  neuro-metasurfaces}.
\newblock \emph{\bibinfo{journal}{Advanced Science}}, \bibinfo{pages}{2204699} (\bibinfo{year}{2021}).

\bibitem{Qian2022}
\bibinfo{author}{Qian, C.} \emph{et~al.}
\newblock \bibinfo{title}{Dynamic recognition and mirage using
  neuro-metamaterials}.
\newblock \emph{\bibinfo{journal}{Nature Communications}}
  \textbf{\bibinfo{volume}{13}}~(1), \bibinfo{pages}{2694}
  (\bibinfo{year}{2022}).

\bibitem{9912362}
\bibinfo{author}{Guo, M.-H.}, \bibinfo{author}{Liu, Z.-N.},
  \bibinfo{author}{Mu, T.-J.} \& \bibinfo{author}{Hu, S.-M.}
\newblock \bibinfo{title}{Beyond self-attention: External attention using two
  linear layers for visual tasks}.
\newblock \emph{\bibinfo{journal}{IEEE Transactions on Pattern Analysis and
  Machine Intelligence}} \textbf{\bibinfo{volume}{45}}~(5),
  \bibinfo{pages}{5436--5447} (\bibinfo{year}{2023}).

\bibitem{doi:10.1021/acs.nanolett.0c01105}
\bibinfo{author}{Schlickriede, C.} \emph{et~al.}
\newblock \bibinfo{title}{Nonlinear imaging with all-dielectric metasurfaces}.
\newblock \emph{\bibinfo{journal}{Nano Letters}}
  \textbf{\bibinfo{volume}{20}}~(6), \bibinfo{pages}{4370--4376}
  (\bibinfo{year}{2020}).

\bibitem{Xiao2018}
\bibinfo{author}{Xiao, Y.}, \bibinfo{author}{Qian, H.} \& \bibinfo{author}{Liu,
  Z.}
\newblock \bibinfo{title}{Nonlinear metasurface based on giant optical kerr
  response of gold quantum wells}.
\newblock \emph{\bibinfo{journal}{ACS Photonics}}
  \textbf{\bibinfo{volume}{5}}~(5), \bibinfo{pages}{1654--1659}
  (\bibinfo{year}{2018}).

\bibitem{8350283}
\bibinfo{author}{Akie, M.}, \bibinfo{author}{Fujisawa, T.},
  \bibinfo{author}{Sato, T.}, \bibinfo{author}{Arai, M.} \&
  \bibinfo{author}{Saitoh, K.}
\newblock \bibinfo{title}{GeSn/SiGeSn multiple-quantum-well electroabsorption
  modulator with taper coupler for mid-infrared Ge-on-Si platform}.
\newblock \emph{\bibinfo{journal}{IEEE Journal of Selected Topics in Quantum
  Electronics}} \textbf{\bibinfo{volume}{24}}~(6), \bibinfo{pages}{1--8}
  (\bibinfo{year}{2018}).

\bibitem{Zuo:19}
\bibinfo{author}{Zuo, Y.} \emph{et~al.}
\newblock \bibinfo{title}{All-optical neural network with nonlinear activation
  functions}.
\newblock \emph{\bibinfo{journal}{Optica}} \textbf{\bibinfo{volume}{6}}~(9),
  \bibinfo{pages}{1132--1137} (\bibinfo{year}{2019}).

\end{thebibliography}

\providecommand{\noopsort}[1]{}\providecommand{\singleletter}[1]{#1}%


\section*{Acknowledgments}

This work was supported by the National Natural Science Foundation of China (Grant Nos. 12064025, 12264028, 12364045 and 12304420); the Natural Science Foundation of Jiangxi Province (Grant Nos. 20212ACB202006, 20232BAB201040 and 20232BAB211025); the Shanghai Pujiang Program (Grant No. 22PJ1402900); the Australian Research Council Discovery Project (Grant No. DP200101353); the Interdisciplinary Innovation Fund of Nanchang University (Grant No. 2019-9166-27060003); the China Scholarship Council (Grant No. 202008420045).

\section*{Author Contributions}

J.Q. and S.X. conceived the idea. J.Q. and T.L. performed the theoretical calculation and numerical simulations. J.Q., T.Y., L.H., and A.M. helped with the theoretical interpretation. J.Q. and D.Z. performed the experiments. T.Y., L.H., and T.L. supervised the project. All authors discussed the results and prepared the manuscript.

\section*{Competing Interests}

The authors declare no competing interests.

\end{document}